\documentclass[sigconf]{acmart}

\usepackage{amsmath,amsfonts}
\usepackage{multicol}
\usepackage{multirow}
\usepackage{bbding}
\usepackage{booktabs}

\AtBeginDocument{%
  \providecommand\BibTeX{{%
    \normalfont B\kern-0.5em{\scshape i\kern-0.25em b}\kern-0.8em\TeX}}}

\copyrightyear{2021}
\acmYear{2021}
\setcopyright{acmlicensed}\acmConference[MM '21]{Proceedings of the 29th ACM International Conference on Multimedia}{October 20--24, 2021}{Virtual Event, China}
\acmBooktitle{Proceedings of the 29th ACM International Conference on Multimedia (MM '21), October 20--24, 2021, Virtual Event, China}
\acmPrice{15.00}
\acmDOI{10.1145/3474085.3478328}
\acmISBN{978-1-4503-8651-7/21/10}

\begin{document}

\title{MMOCR: A Comprehensive Toolbox for Text Detection, Recognition and Understanding}


\author{Zhanghui Kuang}
\affiliation{%
  \institution{SenseTime Research}
  \country{China}
}
\author{Hongbin Sun}
\affiliation{%
  \institution{SenseTime Research}
  \country{China}
}
\author{Zhizhong Li}
\affiliation{%
  \institution{SenseTime Research}
  \country{China}
}
\author{Xiaoyu Yue}
\affiliation{%
  \institution{Center for Perceptual and Interactive Intelligence}
  \country{Hong Kong}
}
\author{Tsui Hin Lin}
\affiliation{%
  \institution{SenseTime Research}
  \country{China}}

\author{Jianyong Chen}
\affiliation{%
  \institution{South China University of Technology}
  \country{China}}

\author{Huaqiang Wei}
\affiliation{%
  \institution{SenseTime Research}
  \country{China}}

\author{Yiqin Zhu}
\affiliation{%
  \institution{South China University of Technology}
  \country{China}}

\author{Tong Gao}
\affiliation{%
  \institution{SenseTime Research}
  \country{China}}

\author{Wenwei Zhang}
\affiliation{%
  \institution{Nanyang Technological University}
  \country{Singapore}}

\author{Kai Chen}
\affiliation{%
  \institution{SenseTime Research, \\ Shanghai AI Laboratory}
  \country{China}}

\author{Wayne Zhang}
\affiliation{%
  \institution{SenseTime Research}
  \country{China}}

\author{Dahua Lin}
\affiliation{%
  \institution{ The Chinese University of Hong Kong}
  \country{Hong Kong}}

\thanks{Jianyong Chen and Yiqin Zhu are students from South China University of
  Technology China.}





\renewcommand{\shortauthors}{Kuang, et al.}

\begin{abstract}
  We present MMOCR---an open-source toolbox which provides a comprehensive pipeline for text detection and recognition, as well as their downstream tasks such as named entity recognition and key information extraction. MMOCR implements 14 state-of-the-art algorithms, which is significantly more than all the existing open-source OCR projects we are aware of to date. To facilitate future research and industrial applications of text recognition-related problems, we also provide a large number of trained models and detailed benchmarks to give insights into the performance of text detection, recognition and understanding. MMOCR is publicly released at \url{https://github.com/open-mmlab/mmocr}.
\end{abstract}






\maketitle

\section{Introduction}

In recent years, deep learning has achieved tremendous success in fundamental computer vision applications such as image recognition~\cite{he2016deep,Simonyan15,Szegedy2014}, object detection~\cite{Girshick2015,ren2015faster,Liu2016,Redmon2016a} and image segmentation~\cite{Long2015,he2017mask}.
In light of this, deep learning has also been applied to areas such as text detection~\cite{Zhu2021,Duan2019,Wang2019a} and text recognition~\cite{Yue2020,li2018show,yang2019symmetry,xie2019aggregation,liao2018scene}, as well as their downstream tasks such as key information extraction~\cite{Sun2021,Yu2020a,Faddoul2018} and named entity recognition~\cite{Chiu2016,Xua}.

Different approaches utilize different training datasets, optimization strategies (\textit{e.g.}, optimizers, learning rate schedules, epoch numbers, pre-trained weights, and data augmentation pipelines), and network designs (\textit{e.g.} network architectures and losses).
To encompass the diversity of components used in various models, we have proposed the MMOCR toolbox which covers recent popular text detection, recognition and understanding approaches in a unified framework.
As of now, the toolbox implements seven text detection methods, five text recognition methods, one key information method and one named entity recognition method.
Integrating various algorithms confers code reusability and therefore dramatically simplifies the implementation of algorithms.
Moreover, the unified framework allows different approaches to be compared against each other fairly and that their key effective components can be easily investigated.
To the best of our knowledge, MMOCR reimplements the largest number of deep learning-based text detection and recognition approaches amongst various open-source toolboxes, and we believe it will facilitate future research on text detection, recognition and understanding.

\begin{figure}[t]
  \begin{minipage}[t]{0.98\linewidth}
    \centering
    \includegraphics[width =1.0\textwidth]{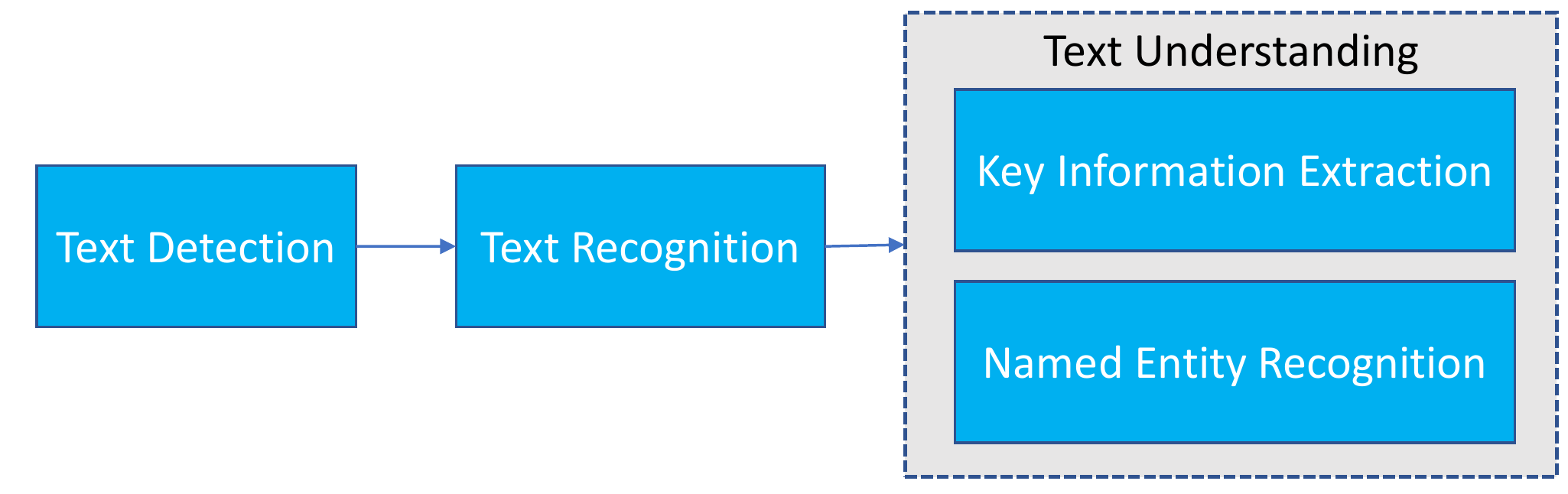}\label{fig:fig_intro_hs}
  \end{minipage}
  \vspace{-8pt}
  \Description[Pipeline of MMOCR]{Text detection, text recognition, and text understanding which includes key information extraction and named entiry recognition}
  \caption{\small
    Overview of MMOCR.
    The supported text detection algorithms include DB~\cite{DBLP:conf/aaai/LiaoWYCB20}, Mask R-CNN~\cite{he2017mask}, PANet~\cite{Wang2019a}, PSENet~\cite{Wang2019}, TextSnake~\cite{DBLP:conf/eccv/LongRZHWY18}, DRRG~\cite{DBLP:conf/cvpr/ZhangZHLYWY20}, and FCENet~\cite{Zhu2021}.
    The supported text recognition algorithms are CRNN~\cite{shi2016end}, NRTR~\cite{sheng2019nrtr}, RobustScanner~\cite{Yue2020}, SAR~\cite{li2018show}, and SegOCR~\cite{segocr}.
    The supported key information extraction algorithm is SDMG-R~\cite{Sun2021}, and the supported named entity extraction algorithm is Bert-Softmax~\cite{Xua}.}\label{fig:fig_intro}
  \vspace{-15pt}
\end{figure}

Extracting structured information such as ``shop name'', ``shop address'' and ``total payment'' in receipt images, and ``name'' and ``organization name'' in document images plays an important role in many practical scenarios.
For example, in the case of office automation, such structured information is useful for efficient archiving or compliance checking.
To provide a comprehensive pipeline for practical applications, MMOCR reimplements not only text detection and text recognition approaches, but also their downstream tasks such as key information extraction and named entity recognition as illustrated in Figure~\ref{fig:fig_intro}.
In this way, MMOCR can meet the document image processing requirements in a one-stop-shopping manner.

MMOCR is publicly released at \url{https://github.com/open-mmlab/mmocr} under the Apache-2.0 License.
The repository contains all the source code and detailed documentation including installation instructions, dataset preparation scripts, API documentation,  model zoo, tutorials and user manual.
MMOCR re-implements more than ten state-of-the-art text detection, recognition, and understanding algorithms, and provides extensive benchmarks and models trained on popular academic datasets.
To support multilingual OCR tasks, MMOCR also releases Chinese text recognition models trained on industrial datasets~\footnote{https://github.com/chineseocr/chineseocr}.
In addition to (distributed) training and testing scripts, MMOCR offers a rich set of utility tools covering visualization, demonstration and deployment.
The models provided by MMOCR are easily converted to onnx~\footnote{https://github.com/onnx/onnx} which is widely supported by deployment frameworks and hardware devices.
Therefore, it is useful for both academic researchers and industrial developers.

\section{Related Work}

\textbf{Text detection.}
Text detection aims to localize the bounding boxes of text instances~\cite{he2017mask,pmtd,DBLP:conf/bmvc/YueKZCH0Z18,Duan2019,Zhu2021,DBLP:conf/mm/WangZQHEHLDS19}.
Recent research focus has shifted to challenging arbitrary-shaped text detection~\cite{Duan2019,Zhu2021}.
While Mask R-CNN~\cite{he2017mask,pmtd} can be used to detect texts,
it might fail to detect curved and dense texts due to the  rectangle-based ROI proposals.
On the other hand, TextSnake~\cite{DBLP:conf/eccv/LongRZHWY18} describes text instances with a series of ordered, overlapping disks.
PSENet~\cite{Wang2019} proposes a progressive scale expansion network which enables the differentiation of curved text instances that are located close together.
DB~\cite{DBLP:conf/aaai/LiaoWYCB20} simplifies the post-processing of binarization for scene-text segmentation by proposing a differentiable binarization function to a segmentation network, where the threshold value at every point of the probability map of an image can be adaptively predicted.

\textbf{Text recognition.}
Text recognition has gained increasing attention due to its ability to extract rich semantic information from text images.
Convolutional Recurrent Neural Network (CRNN)~\cite{shi2016end} uses an end-to-end trainable neural network which consists of a Deep Convolutional Neural Networks (DCNN) for the feature extraction, a Recurrent Neural Networks (RNN) for the sequential prediction and a transcription layer to produce a label sequence.
RobustScanner~\cite{Yue2020} is capable of recognizing contextless texts by using a novel position enhancement branch and a dynamic fusion module which mitigate the misrecognition issue of random text images.
Efforts have been made to rectify irregular texts input into regular ones which are compatible with typical text recognizers.
For instance, Thin-Plate-Spline (TPS) transformation  is employed in a deep neural network that combines a Spatial Transformer Network (STN) and a Sequence Recognition Network (SRN) to rectify curved and perspective texts in STN before they are fed into SRN~\cite{shi2018aster}.

\textbf{Key information extraction.}
Key Information Extraction (KIE) for unstructured document images, such as receipts or credit notes, is most notably used for office automation tasks including efficient archiving and compliance checking.
Conventional approaches, such as template matching, fail to generalize well on documents of unseen templates.
Several models are proposed to resolve the generalization problem.
For example, CloudScan~\cite{Palm2017} employs NER to analyze the concatenated one-dimensional text sequence for the entire invoice.
Chargrid~\cite{Faddoul2018} encodes each document page as a two-dimensional grid of characters to conduct semantic segmentation, but it cannot make full use of the non-local, distant spatial relation between text regions since it covers two-dimensional spatial layout information with small neighborhood only.
Recently, an end-to-end Spatial Dual Modality Graph Reasoning (SDMG-R) model~\cite{Sun2021} has been developed which is particularly robust against text recognition errors.
It models unstructured document images as spatial dual-modality graphs with graph nodes as detected text boxes and graph edges as spatial relations between nodes.

\textbf{Named entity recognition.}
Named entity recognition (NER)~\cite{Xua,Chiu2016,lample2016neural,zheng-etal-2017-joint} aims to locate and classify named entities into pre-defined categories such as the name of a person or organization.
They are based on either bidirectional LSTMs or conditional random fields.

\begin{table*}[t]
  \footnotesize
  \begin{center}
    \caption{Comparison between different open-source OCR toolboxes.} \label{tab:comp}
    \vspace{-1.em}
    \begin{tabular}{c|cccccc}
      \toprule
      Toolbox & tesseract & chineseocr & chineseocr\_lite & EasyOCR & PaddleOCR & MMOCR \\
      \midrule
      DL library & --- & PyTorch & PyTorch & PyTorch & PaddlePaddle & PyTorch \\
      \multirow{2}{*}{Inference engine} & --- & OpenCV DNN & NCNN & PyTorch & Paddle inference & PyTorch \\
       & & & TNN & & Paddle lite & onnx runtime \\
       & & & onnx runtime & & & TensorRT \\
      \multirow{4}{*}{OS} & --- & Windows & Windows & Windows & Windows & Windows \\
       & Linux & Linux & Linux & Linux & Linux & Linux \\
       & Android & --- & Android & --- & Android & --- \\
       & IOS & --- & IOS & --- & IOS & --- \\
       Language\# & 100+ & 2 & 2 & 80+ & 80+ & 2 \\
      \multirow{2}{*}{Detection} & convention & YOLOV3~\cite{Redmon2018} & DB~\cite{DBLP:conf/aaai/LiaoWYCB20} & CRAFT~\cite{Baek2019} & EAST~\cite{Zhou2017}, DB~\cite{DBLP:conf/aaai/LiaoWYCB20}, SAST~\cite{DBLP:conf/mm/WangZQHEHLDS19} & MaskRCNN~\cite{he2017mask}, PAN~\cite{Wang2019a}, PSENet~\cite{Wang2019} \\
      & & & & & & DB~\cite{DBLP:conf/aaai/LiaoWYCB20}, TextSnake~\cite{DBLP:conf/eccv/LongRZHWY18}, DRRG~\cite{DBLP:conf/cvpr/ZhangZHLYWY20}, FCENet~\cite{Zhu2021} \\
      \multirow{2}{*}{Recognition} & convention & CRNN~\cite{shi2016end} & DB~\cite{DBLP:conf/aaai/LiaoWYCB20} & CRNN~\cite{shi2016end} & CRNN~\cite{shi2016end}, Rosetta~\cite{DBLP:journals/corr/abs-1910-05085}, SRN~\cite{DBLP:conf/cvpr/YuLZLHLD20} & CRNN~\cite{shi2016end}, RobustScanner~\cite{Yue2020}, SAR~\cite{li2018show} \\
      & LSTM & & & & Star-Net~\cite{liu2016star}, RARE~\cite{shi2016robust} & SegOCR~\cite{segocr}, Transformer~\cite{li2018show} \\
      Downstream tasks & & & & & & KIE, NER \\
      Support training & Yes & Yes & No & No & Yes & Yes \\
      \bottomrule
    \end{tabular}
  \end{center}
\end{table*}

\begin{table*}
  \footnotesize
  \caption{The effects of backbones.
    All models are pre-trained on ImageNet, and trained on ICDAR2015 training set and evaluated on its test set.}
  \begin{center}
    \vspace{-1.em}
    \begin{tabular}{ c|c| c c c|c c c| c c c  }
      \toprule
      & & \multicolumn{3}{c|}{PSENet} & \multicolumn{3}{c|}{PAN} & \multicolumn{3}{c}{DB} \\
      Backbone & FLOPs & Recall & Precision & H-mean & Recall & Precision & H-mean & Recall & Precision & H-mean \\ \midrule
      ResNet18 & 37.1G & 73.5 & 83.8 & 78.3 & 73.4 & 85.6 & 79.1 & 73.1 & 87.1 & 79.5 \\
      ResNet50 & 78.9G & 78.4 & 83.1 & 80.7 & 73.2 & 85.5 & 78.9 & 77.8 & 82.1 & 79.9 \\
      ddrnet23-slim~\cite{DBLP:journals/corr/abs-2101-06085} & 16.7G & 75.2 & 80.1 & 77.6 & 72.3 & 83.4 & 77.5 & 76.7 & 78.5 & 77.6 \\ \bottomrule
    \end{tabular}
  \end{center}
  \label{tab:comp_backbone}
\end{table*}

\begin{table*}
  \footnotesize
  \begin{center}
    \caption{The effects of necks.
      All models are pre-trained on ImageNet, and trained on ICDAR2015 training set and evaluated on its test set.}\label{tab:comp_neck}
    \vspace{-1.em}
    \begin{tabular}{ c|c| c c c|c c c| c c c  }
      \toprule
      & & \multicolumn{3}{c|}{PSENet} & \multicolumn{3}{c|}{PAN} & \multicolumn{3}{c}{DB} \\ \midrule
      Necks & FLOPs  & Recall & Precision & H-mean & Recall & Precision & H-mean & Recall & Precision & H-mean \\
      FPNF~\cite{Wang2019} & 208.6G & 78.4 & 83.1 & 80.7 & 72.4 & 86.4 & 78.8 & 77.5 & 82.3 & 79.8   \\
      PFNC~\cite{DBLP:conf/aaai/LiaoWYCB20} & 22.4G  & 75.6 & 80.0 & 77.7 & 70.9 & 83.3 & 76.6 & 73.1 & 87.1 & 79.5 \\
      FPEM\_FFM~\cite{Wang2019a} & 7.79G & 71.7 & 82.0 & 76.5 & 73.4 & 85.6 & 79.1 & 71.8 & 86.7 & 78.6   \\ \bottomrule
    \end{tabular}
  \end{center}
\end{table*}

\textbf{Open source OCR toolbox.}
Several open-source OCR toolboxes have been developed over the years to meet the increasing demand from both academia and industry.
Tesseract\footnote{\url{https://github.com/tesseract-ocr/tesseract}} is the pioneer of open-source OCR toolbox.
It was publicly released in 2005, and provides CLI tools to extract printed font texts from images.
It initially followed a traditional, step-by-step pipeline comprising the connected component analysis, text line finding, baseline fitting, fixed pitch detection and chopping, proportional word finding, and word recognition~\cite{Smith2007}.
It now supports an LSTM-based OCR engine and supports more than 100 languages.
Deep learning-based open-source OCR toolbox EasyOCR~\footnote{\url{https://github.com/JaidedAI/EasyOCR}} has been released recently.
It provides simple APIs for industrial users and supports more than 80 languages.
It implemented the CRAFT~\cite{Baek2019} detector and CRNN~\cite{shi2016end} recognizer.
However, it is for inference only and does not support model training.
ChineseOCR~\footnote{\url{https://github.com/chineseocr/chineseocr}} is another popular open-source OCR toolbox.
It uses YOLO-v3~\cite{Redmon2018} and CRNN~\cite{shi2016end} for text detection and recognition respectively, and uses OpenCV DNN for deep models inference.
By contrast, ChineseOCR\_lite~\footnote{\url{https://github.com/DayBreak-u/chineseocr\_lite}} releases a lightweight Chinese detection and recognition toolbox that uses DB~\cite{DBLP:conf/aaai/LiaoWYCB20} to detect texts and CRNN~\cite{shi2016end} to recognize texts.
It provides forward inference based on NCNN~\footnote{\url{https://github.com/Tencent/ncnn}} and TNN~\footnote{\url{https://github.com/Tencent/TNN}}, and can be deployed easily on multiple platforms such as Windows, Linux and Android.
PaddleOCR~\footnote{\url{https://github.com/PaddlePaddle/PaddleOCR}} is a practical open-source OCR toolbox based on PaddlePaddle and can be deployed on multiple platforms such as Linux, Windows and MacOS.
It currently supports more than 80 languages and implements three text detection methods (EAST~\cite{Zhou2017}, DB~\cite{DBLP:conf/aaai/LiaoWYCB20}, and SAST~\cite{DBLP:conf/mm/WangZQHEHLDS19}), five recognition methods (CRNN~\cite{shi2016end}, Rosetta~\cite{DBLP:journals/corr/abs-1910-05085}, STAR-Net~\cite{liu2016star}, RARE~\cite{shi2016robust} and SRN~\cite{DBLP:conf/cvpr/YuLZLHLD20}), and one end-to-end text spotting method (PGNet)~\cite{DBLP:conf/aaai/WangZQLZLHLDS21}.
Comprehensive comparisons among these open-source toolboxes are given in Table~\ref{tab:comp}.

\section{Text Detection Studies}
Many important factors can affect the performance of deep learning-based models.
In this section, we investigate the backbones and necks of network architectures.
We exchange the above components between different segmentation-based text detection approaches to measure the performance and computational complexity effects.

\textbf{Backbone.} ResNet18~\cite{he2016deep} and ResNet50~\cite{he2016deep} are commonly used in text detection approaches.
For practical applications, we also introduce a GPU-friendly lightweight backbone ddrnet23-slim~\cite{DBLP:journals/corr/abs-2101-06085}.
Table~\ref{tab:comp_backbone} compares ResNet18, ResNet50 and ddrnet23-slim in terms of FLOPs and H-mean by plugging them in PSENet, PAN and DB.
It has been shown that ddrnet23-slim performs slightly worse than ResNet18 and ResNet50, as it only has 45\% and 21\% FLOPs of ResNet18 and ResNet50 respectively.

\textbf{Neck.} PSENet, PAN and DB propose different FPN-like necks to fuse multi-scale features.
Our experimental results in Table~\ref{tab:comp_neck} show that the FPNF proposed in PSENet~\cite{Wang2019} can achieve the best H-mean in PSENet and DB~\cite{DBLP:conf/aaai/LiaoWYCB20}.
However, its FLOPs are substantially higher than those of PFNC proposed in DB~\cite{DBLP:conf/aaai/LiaoWYCB20} and FPEM\_FFM proposed in PAN~\cite{Wang2019a}.
By contrast, FPEM\_FFM has the lowest FLOPs and achieves the best H-mean in PAN~\cite{Wang2019a}.

\section{Conclusions}
We have publicly released MMOCR, which is a comprehensive toolbox for text detection, recognition and understanding. MMOCR has implemented 14 state-of-the-art algorithms, which is more than all the existing open-source OCR projects. Moreover, it has offered a wide range of  trained models, benchmarks, detailed documents, and utility tools. In this report, we have extensively compared MMOCR with other open-source OCR projects. Besides, we have introduced a GPU-friendly lightweight backbone-ddrnet23-slim, and carefully studied the effects of backbones and necks in terms of detection performance and computational complexity which can guide  industrial applications.

\section{Acknowledgement}
This work was supported by the Shanghai Committee of Science and Technology, China (Grant No. 20DZ1100800).

\bibliographystyle{ACM-Reference-Format}
\bibliography{sample-base}


\end{document}